\pdfoutput=1

\documentclass[11pt]{article}

\usepackage[]{EMNLP2023}

\usepackage{times}
\usepackage{latexsym}

\usepackage[T1]{fontenc}

\usepackage[utf8]{inputenc}

\usepackage{microtype}

\usepackage{inconsolata}
\usepackage{booktabs}
\usepackage{multirow}
\usepackage{amsmath, amssymb}
\usepackage{graphicx}
\usepackage{algorithm}  
\usepackage[noend]{algorithmic}

\renewcommand{\algorithmiccomment}[1]{\bgroup\hfill//~#1\egroup}

\title{RTF: Region-based Table Filling Method for Relational Triple Extraction}

\author{
    \textbf{Ning An},
    \textbf{Lei Hei}, 
    \textbf{Yong Jiang}, 
    \textbf{Weiping Meng}, 
    \textbf{Jingjing Hu} \\
    \textbf{Boran Huang},
    \textbf{Feiliang Ren}\thanks{\quad Corresponding author.} \\
    School of Computer Science and Engineering \\
    Northeastern University, Shenyang, 110169, China \\
    \texttt{renfeiliang@cse.neu.edu.cn}
}

\begin{document}
\maketitle
\begin{abstract}

Relational triple extraction is crucial work for the automatic construction of knowledge graphs. Existing methods only construct shallow representations from a token or token pair-level. However, previous works ignore local spatial dependencies of relational triples, resulting in a weakness of entity pair boundary detection. To tackle this problem, we propose a novel \textbf{R}egion-based \textbf{T}able \textbf{F}illing method (RTF). We devise a novel region-based tagging scheme and bi-directional decoding strategy, which regard each relational triple as a region on the relation-specific table, and identifies triples by determining two endpoints of each region. We also introduce convolution to construct region-level table representations from a spatial perspective which makes triples easier to be captured. In addition, we share partial tagging scores among different relations to improve learning efficiency of relation classifier. Experimental results show that our method achieves state-of-the-art with better generalization capability on three variants of two widely used benchmark datasets.

\end{abstract}

\section{Introduction}

Relational Triple Extraction (RTE) aims to extract relation triples with form $\textit{(subject, relation, object)}$ from massive unstructured text. This task is one of the most fundamental tasks in information extraction, thus can be widely applied to various downstream applications in the real world~\cite{DBLP:journals/cogcom/NayakMGP21}.

Early works~\cite{DBLP:conf/emnlp/ZelenkoAR02,DBLP:conf/acl/ZhouSZZ05, DBLP:conf/acl/ChanR11} often decompose RTE task into two consecutive sub-tasks, entity extraction and relation classification of entity pairs. This pipeline methods suffer from error propagation problems due to inconsistencies during the training and inference stages~\cite{DBLP:conf/acl/LiJ14}.

Therefore, end-to-end joint extraction methods have received more widespread attention in recent years~\citep{DBLP:conf/coling/GuptaSA16, DBLP:conf/acl/MiwaB16, DBLP:conf/acl/ZhengWBHZX17, DBLP:conf/emnlp/ZhangZF17}. Joint extraction aims to extract both entity pairs and their relations simultaneously. The table filling-based approach is one of the best-performing methods for joint extraction. This approach converts RTE into sequence labeling by creating a relation-specific table for each relation and identifying subject and object pairs on the table separately. However, previous RTE methods only construct shallow features at the token or token pair-level, such as applying linear transformations to a single token or token pair. These approaches fail to capture the local spatial dependencies of the relational triples, thus neither allowing token pairs to interact explicitly over a two-dimensional structure, nor modeling interactions within and across triples. This may lead to weaker detection of entity pair boundaries. 


\begin{figure}[t]
    \centering
    \includegraphics[width=0.35\textwidth]{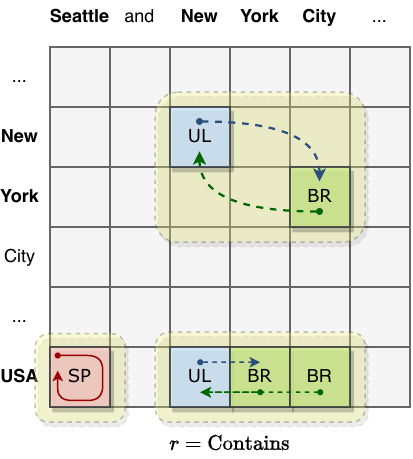}
    \caption{Example of our EPR tagging scheme and bi-directional decoding. This sentence contains Single Entity Overlapping (SEO) and Head Tail Overlapping (HTO) overlapping patterns. }
    \label{fig:TaggingStrategy}
\end{figure}

As we have observed, each relational triple corresponds to a fixed rectangular region on a relation-specific table. If we can identify the boundaries of entity pairs from a spatial perspective, we can extract triples more conveniently. For example, consider the input sentence ``\textit{Seattle and New York City in New York are popular in the USA.}'' in Figure~\ref{fig:TaggingStrategy}. Suppose the relation of the table is ``$\textit{Contains}$'', and there are four $(\textit{subject},\textit{object})$ pairs: $(\textit{New York, New York City})$, $(\textit{USA, Seattle})$, $(\textit{USA, New York})$, and $(\textit{USA, New York City})$. Figure~\ref{fig:TaggingStrategy} represents a wealth of regional information that has been overlooked by other works\citep{DBLP:conf/coling/WangYZLZS20, DBLP:conf/aaai/ShangHM22}. This figure clearly illustrates that each entity pair can be treated as a rectangular region, which can be determined by recognizing the upper-left and bottom-right endpoints. One special case is $(\textit{USA}, \textit{Seattle})$, where the rectangle degenerates to a single point. Furthermore, entity pair boundary divisions are more explicit from a region-level perspective. Such as SEO entity pairs $(\textit{USA, New York})$ and $(\textit{USA, New York City})$, when the tag of $(\textit{USA}, \textit{New})$ depends not only on the token pair itself but also on its surrounding region, it is easier to argue that $\textit{USA}$ plays a subject role in some relational triple. Overlapping triples are also more likely to be detected at once during this process. 

Inspired by the above observations, we propose a simple but effective \textbf{R}egion-based \textbf{T}able-\textbf{F}illing (RTF) method to handle neglecting spatially local dependencies when extracting relational triples. We devise a novel \textbf{E}ntity \textbf{P}air as \textbf{R}egion (EPR) tagging scheme and bi-directional decoding strategy to identify each entity pair on the table. 
EPR tagging scheme recognizes entity pairs by identifying the upper-left and lower-right corners of the region, which can also handle various types of overlapping triples. Bi-directional decoding provides some fault tolerance to the model to a certain extent. To exploit the spatial local dependencies, we apply convolution to construct region-level table representations. This allows each token pair representation not only to depend on itself but also on the surrounding token pairs. When the convolutional kernel slides over the table, the entire triple or even across triples can be simultaneously interacted. Moreover, we shared a portion of tagging scores among different relations, i.e., the scores are divided into relation-dependent and relation-independent parts. This allows the relation classifier to learn the relation-specific residuals only, which reduces the burden of the relation classifier.

We comprehensively conducted experiments on two benchmark datasets, NYT~\cite{DBLP:conf/pkdd/RiedelYM10} and WebNLG~\cite{gardent2017creating}. Experimental results show that our approach achieves state-of-the-art performance with better generalization. We summarize our contributions as follows:
\begin{itemize}
    \item We propose a novel region-based table filling relational triple extraction method, which devises a novel EPR tagging scheme and bi-directional decoding algorithm to extract triples by determining the upper-left and bottom-right endpoints of the region.
    \item We introduce convolution to construct region-level table representations and share tagging scores of different relations to fully utilize regional correlations for improving entity pair boundary recognition.
    \item Our approach achieves state-of-the-art on two benchmark on three variants of two widely used benchmark datasets with superior generalization capability.
\end{itemize}

\section{Related Work}

Early works~\cite{DBLP:conf/emnlp/ZelenkoAR02,DBLP:conf/acl/ZhouSZZ05, DBLP:conf/acl/ChanR11} decomposes the relational triple extraction into two separate sub-tasks, entity recognition and relation classification between entity pairs. In this paradigm, all the entities in a sentence are first extracted, and then the extracted entities are combined into entity pairs to determine the relations between them. This pipeline approach suffers from error propagation~\cite{DBLP:conf/acl/LiJ14} due to inconsistency in training and inference.

To eliminate inconsistencies during the training and inference phases, joint entity and relation extraction~\citep{DBLP:conf/coling/GuptaSA16, DBLP:conf/acl/MiwaB16, DBLP:conf/acl/ZhengWBHZX17, DBLP:conf/emnlp/ZhangZF17} is more widely researched. We roughly divide the existing research methods into three categories according to the research route.

\noindent \textbf{Sequence Tagging Methods}
~\citet{DBLP:conf/acl/WeiSWTC20} proposes a cascade framework that first extracts all subjects in a sentence and then predicts the corresponding object under each relation.~\citet{DBLP:conf/acl/ZhengWCYZZZQMZ20} predicts potential relations in a sentence and aligns extracted subject and object pairs in a one-dimensional sequence with a two-dimensional matrix.~\citet{DBLP:conf/emnlp/LiLDYLH21} used translating decoding to accomplish one-dimensional sequence tagging, similar to TransE~\cite{DBLP:conf/nips/BordesUGWY13}.

\noindent \textbf{Table Filling Methods} Recently, table filling-based methods have attracted widespread attention from researchers due to their superior performance.~\citet{DBLP:conf/coling/WangYZLZS20} models entity pair extraction as a token pair linking problem on two-dimensional tables, eliminating gaps in training and inference.~\citet{DBLP:conf/emnlp/YanZFZW21} proposes a two-way interaction method for balanced information transmission from two-task interaction perspective between named entity recognition and relation extraction.~\citet{DBLP:conf/aaai/ShangHM22} extracts entity pairs using the relation-specific horn tagging scheme and resolves RTE task by fine-grained triple classification.

\noindent \textbf{Sequence to Sequence Methods} Some works use encoder-decoder framework to copy or generate relational triples.~\citet{DBLP:conf/acl/LiuZZHZ18,DBLP:conf/emnlp/ZengHZLLZ19,DBLP:conf/aaai/ZengZL20} applies the copy mechanism to copy each triple element from the sentence.~\citet{DBLP:journals/corr/abs-2011-01675} regards RTE as a set prediction problem of relational triples, which avoids exposure bias in generative methods.~\citet{DBLP:conf/emnlp/CabotN21} extracts each triple via pre-training and structured generation.~\citet{DBLP:conf/emnlp/Tan0HZC0Z22} focusing on constructing instance-level representations for relational triples.

However, either type of method ignores the spatial local dependencies of triples. Even though table filling-based RTE methods have been constructed with two-dimensional table representations, tagging is only performed with a token pair-level representation. Compared with existing works, our goal is to construct region-level table representations with regional correlations on two-dimensional table structures, to tackle the problem of underutilized local spatial dependencies of triples.

\section{Methodology}

\subsection{Task Definition}

For a given sentence $S=\left(x_1, x_2, \dots, x_{N}\right)$ containing $N$ tokens, the target of the relational triple extraction task is to extract all the triples $T=\{(e_i, r_t, e_j) | e \in \mathcal{E}, r \in \mathcal{R}\}$, where $e_i$, $e_j$, $r_t$ are subject, object and the relation between them respectively, $\mathcal{E}$ is the set of entities formed by consecutive tokens, and $r_t$ belongs to the predefined set of relations $\mathcal{R}=\{r_1, r_2, \dots, r_m\}$.

Identifying the entity pair $(e_i, e_j)$ can be regarded as finding the start and end positions $(x_i, x_j)$ of $e_i$, and identifying the start and end positions $(x_p, x_q)$ of $e_j$. Since the boundaries of $(e_i, e_j)$ are determined simultaneously on each relation-specific table, entity pairs and relations can be extracted in parallel.

\subsection{EPR Tagging Scheme}

We propose a novel \textbf{E}ntity \textbf{P}air as \textbf{R}egion (EPR) tagging scheme, which can identify triples from the spatial perspective. Following previous table filling-based approaches~\citep{DBLP:conf/coling/WangYZLZS20, DBLP:conf/aaai/ShangHM22}, we maintain a relation-specific two-dimensional table for each relation $r$, and identify entity pairs $(e_i, e_j)$ on each table. 


Specifically, entity pairs $(e_i, e_j)$ need to be determined by the start and end positions $(x_i, x_j)$ of subject $e_i$ and the start and end positions $(x_p, x_q)$ of object $e_j$. Obviously, this is equivalent to finding the \textbf{U}pper-\textbf{L}eft coordinate $(x_i, x_p)$ and the \textbf{B}ottom-\textbf{R}ight coordinate $(x_j, x_q)$ of a rectangular region on a relation-specific table, which we denote by ``\textbf{UL}'' and ``\textbf{BR}'', respectively. When the subject and object both have a single token, the entity pair coincides with the special case of a rectangular region collapsed into a \textbf{S}ingle \textbf{P}oint, which we represent with ``\textbf{SP}'' individually. All remaining cells are tagged as ``None''. As shown in Figure~\ref{fig:TaggingStrategy}, entity pair $(\textit{New York}, \textit{New York City})$ is a rectangular region on the relation ``$\textit{Contains}$'' table, and the upper-left corner of the region $(\textit{New}, \textit{New})$ and the bottom-right corner $(\textit{York, City})$ should be filled with ``UL'' and ``BR'' respectively. The corresponding decoding strategy is described in Section~\ref{sec:decoding}.

Our tagging scheme can address various overlapping patterns in complex scenarios~\citep{DBLP:conf/acl/LiuZZHZ18}, such as Single Entity Overlapping (SEO), Entity Pair Overlapping (EPO), Head Tail Overlapping (HTO), because they are all non-intersecting in regional perspective. Notably, although our tagging scheme can be similar to OneRel~\cite{DBLP:conf/aaai/ShangHM22}, it takes up 3 times more memory than EPR because OneRel ignores the case where a triple degenerates to a single point. 



\subsection{RTF}

\begin{figure*}
    \centering
    \includegraphics[width=1\textwidth]{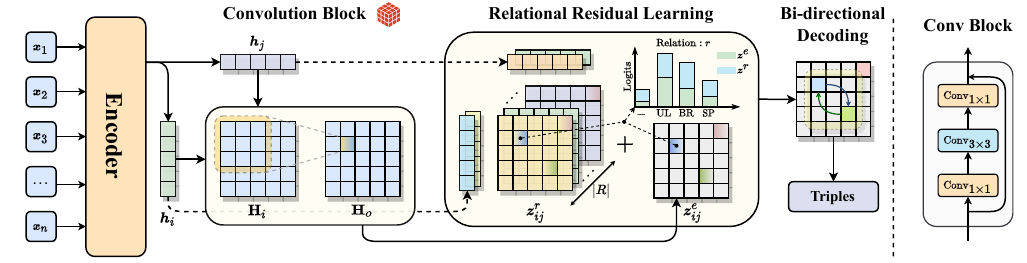}
    \caption{RTF overview. \textbf{Left}: First, we construct a token pair-level table representation. Then capture the local dependencies and construct a region level table representation by Convolution Block. After that, we obtain the tagging score of each cell from the sharing score of all relations $\boldsymbol{z}_{ij}^e$ and the relation-specific residual $\boldsymbol{z}_{ij}^r$. Finally, we obtain triples by bi-directional decoding strategy. \textbf{Right}: Details of the Convolution Block. Similar to ResNet~\cite{DBLP:conf/cvpr/HeZRS16}, Conv Block consists of $1\times1, 3\times3, 1\times1$ convolution and residual connection.}
    \label{fig:model}
\end{figure*}

\subsubsection{Table Representation}

Given an input sentence $S = \left(x_1, x_2, \dots, x_N\right)$, we first use the pre-trained language model to obtain the contextual representation of each token $x_i$ in the sentence:

\begin{equation}
\boldsymbol{h}_1, \boldsymbol{h}_2, \dots, \boldsymbol{h}_N = \text{PLM} \left(x_1, x_2, \dots, x_N \right)
\end{equation}

where $\boldsymbol{h}_i \in \mathbb{R}^{d}$ is token contextual representation, $N$ is the sequence length of sentence. $\rm{PLM}$ can be arbitrary pre-trained bi-directional language models, i.e. BERT~\cite{DBLP:conf/naacl/DevlinCLT19}. 

Then we use a linear transformation to construct the shallow representation of each token pair $(x_i, x_j)$ in the two-dimensional table:

\begin{equation}
\boldsymbol{h}_{ij} = \phi\left(\boldsymbol{W}_t \left[\boldsymbol{h}_i; \boldsymbol{h}_j\right] + \boldsymbol{b}_t\right)
\end{equation}

where $\phi(\cdot)$ denotes \rm{ReLU} non-linear activation function, $[;]$ is the concatenation operation, $\boldsymbol{W}_t \in \mathbb{R}^{2d \times d}, \boldsymbol{b}_t \in \mathbb{R}^{d}$ are trainable parameters.

\subsubsection{Convolution Block}

To construct enriched region-level table representations, we need a component to fully exploit the association of triples over regions, which can be useful for detecting entity pairs. We think that convolution is very appropriate to achieve this purpose. For instance, in Figure~\ref{fig:TaggingStrategy}, when we take a $3 \times 3$ convolution kernel to construct a region-level table representation, the central token pair can interact with 8 surrounding token pairs explicitly. Convolution can perceive the difference in semantic information inside (denoted by yellow rectangle) and outside (indicated by the gray cells outside the yellow rectangle) the entity pair boundary\footnote{We illustrate this process in Appendix~\ref{sec:conv_interact}.}. When the convolution kernel covers the bottom yellow region of the table, it not only allows the entire region of $(\textit{USA}, \textit{New York})$ to interact simultaneously, but also allows $(\textit{USA}, \textit{New York})$, $(\textit{USA}, \textit{New York City})$ to interact across triples on the table.


ResNet~\cite{DBLP:conf/cvpr/HeZRS16} has been proven to be an effective convolutional neural network backbone in computer vision, which is especially strong at capturing local dependencies from a spatial perspective. Therefore, we adopt the Convolution Block as our component for constructing region-level table representations. Token pair-level table representation $\mathbf{H}_i$ is composed of $\boldsymbol{h}_{ij}$. When $\mathbf{H}_i$ is input into the Convolution Block, the region-level table representation $\mathbf{H}_o$ will be obtained. So our Convolution Block is defined as follows:

\begin{equation}
\begin{aligned}
\mathbf{H}_o & = \phi\left(\mathrm{Conv}_{1\times 1}\left(\mathbf{H}_i\right)\right) \\
\mathbf{H}_o & = \phi\left(\mathrm{Conv}_{3\times 3}\left(\mathbf{H}_o\right)\right) \\
\mathbf{H}_o & = \phi\left(\mathrm{Conv}_{1\times 1}\left(\mathbf{H}_o\right)\right) \\
\mathbf{H}_o & = \mathbf{H}_o + \mathbf{H}_i
\end{aligned}
\end{equation}

where $\mathbf{H} \in \mathbb{R}^{N \times N \times d}$ is table representation. Furthermore, layer normalization~\cite{DBLP:journals/corr/BaKH16} is applied after each convolution operation to accelerate the convergence of the trainable parameters in the convolution layer.

\subsubsection{Relational Residual Learning}


Regardless of the relation categories, the ability to determine entity pairs by finding regions on each table should be generalized across all relations, rather than restricted to particular relations. For instance, in Figure~\ref{fig:TaggingStrategy}, $(\textit{New}, \textit{New})$ looks like it could be the head of some entity pair (i.e., the upper-left corner of the region), so it should be slightly more probable to fill with ``UL'' under any relation. In particular, a large number of negative cells on each table should be filled with ``None'' in either relation.


Inspired by~\citet{DBLP:journals/corr/abs-2208-03054}, we separate the tagging score of each cell $\boldsymbol{z}_{ij}$ into the sum of two parts: relation-independent score $\boldsymbol{z}_{ij}^{e}$ and relation-dependent score $\boldsymbol{z}_{ij}^r$. As shown in Figure~\ref{fig:model}, $\boldsymbol{z}_{ij}^{e}$ is considered as a basic score for all cells, which is obtained from the region-level table representation. $\boldsymbol{z}_{ij}^r$ is regarded as the relation-specific residual based on $\boldsymbol{z}_{ij}^{e}$, which learned by different relation classifiers. This means that each relation classifier only needs to focus on fitting the residual $\boldsymbol{z}_{ij} - \boldsymbol{z}_{ij}^{e}$ for a particular relation, rather than on fitting $\boldsymbol{z}_{ij}^e$, because $\boldsymbol{z}_{ij}^e$ is handled by one additional classifier which shared by all relations.




Therefore, if all classifiers can share and leverage this relation-independent information, not only can alleviate the severe class imbalance with the introduction of convolution~\cite{DBLP:journals/nn/BudaMM18}, but also help alleviate the learning burden of relation-specific classifiers.

Specifically, under the relation $r$, the logits $\boldsymbol{z}_{ij}$ of token pair $(x_i, x_j)$ in the table are divided into $\boldsymbol{z}_{ij}^{e}$ and $\boldsymbol{z}_{ij}^{r}$, which are computed as follows:

\begin{equation}
    \begin{aligned}
        \boldsymbol{z}_{ij}^{e} &= \boldsymbol{W}_{e} \boldsymbol{h}_{ij} + \boldsymbol{b}_{e} \\
        \boldsymbol{z}_{ij}^{r} &= \boldsymbol{W}_{r} \left[\boldsymbol{h}_i; \boldsymbol{h}_j\right] + \boldsymbol{b}_{r}
    \end{aligned}
\end{equation}

where $\boldsymbol{W}_e, \boldsymbol{W}_r \in \mathbb{R}^{d \times |L|}, \boldsymbol{b}_e, \boldsymbol{b}_r \in \mathbb{R}^{|L|}$ are trainable parameters. $|L|$ is the number of tags, which is 4 under our EPR tagging scheme.

We simply add these two scores directly and use Softmax to obtain the probability distribution $P_r$ of each tag in each cell:

\begin{equation}
P_r(\mathbf{y}_{ij} | S) = \mathrm{Softmax}\left(\boldsymbol{z}_{ij}^{e} +\boldsymbol{z}_{ij}^{r}\right)
\end{equation}

\subsection{Bi-directional Decoding}\label{sec:decoding}

Our decoding process is logically straightforward. Since two points determine a region (or a triple), only a heuristic nearest neighbor matching algorithm is required to match ``UL'' to the nearest ``BR'' on the bottom-right, and similarly, to match ``BR'' to the nearest ``UL'' on the upper-left. However, the model may give wrong predictions, so the decoding algorithm is required to have some fault tolerance mechanism. For example, the model may misjudge the number of tokens of an entity, which may result in ``UL'' or ``BR'' being identified as ``SP''. In this case, when the model ``UL''/``BR'' fails to match ``BR''/``UL'', we should try to match ``SP''. Our pseudo-code of bi-directional decoding algorithm is presented in Algorithm~\ref{alg:decoding}.

\begin{algorithm}[htbp]
	\footnotesize
	\algsetup{
		linenodelimiter = {  }
	}
    \caption{Bi-directional Decoding Strategy}
    \label{alg:decoding}
	\begin{algorithmic}[1]
		\REQUIRE Given sentence $S$ = ($x_1$, $x_2$, ..., $x_N$), predefined relation set $\mathcal{R}$, prediction labels $\mathbf{Y}_r \in \mathbb{R}^{N \times N}$ for the 2D table of each relation $r \in \mathcal{R}$. \\
		\ENSURE Predicted relational triple set $T$.
		\FOR{each $r \in \mathcal{R}$}
		\STATE 
            Find the token pairs predicted as ``SP'', ``UL'', ``BR'' from the relation-specific table $\mathbf{Y}_r$, denoted as $\mathbf{Y}_r^{\textit{SP}}$, $\mathbf{Y}_r^{\textit{UL}}$ and $\mathbf{Y}_r^{\textit{BR}}$, respectively.
  
		\FOR{token pair $(x_i,x_j) \in \mathbf{Y}_r^{\textit{SP}}$} 
		\STATE 
            Extract entity pair $(s, o)$ from sentence $S$ under $(x_i, x_j)$, then add triple $(s, r, o)$ to $T$. 
		\ENDFOR		
		
		\FOR[UL $\rightarrow$ BR]{token pair $(x_i, x_j) \in \mathbf{Y}_r^{\textit{UL}}$}
        \STATE 
            Compute the Euclidean distance between each token pair $(x_i, x_j)$ and the token pair $(x_m, x_n) \in \mathbf{Y}_r^{\textit{BR}}$ satisfying the token pair located in the bottom-right, trying to match with the nearest ``BR''.
                
        \IF{Nearest ``BR'' $(x_{p}, x_{q})$ \textbf{not} exists}
        \STATE
            Compute the Euclidean distance between each token pair $(x_i, x_j)$ and the token pair $(x_m, x_n) \in \mathbf{Y}_r^{\textit{SP}}$ satisfying the token pair located in the bottom-right, trying to match with the nearest ``SP''.
            
        \ENDIF
        \IF{Nearest location $(x_{p}, x_{q})$ exists}
        \STATE 
            Extract the entity pair $(s, o)$ from the sentence $S$ according to subject position $(x_i, x_p)$ and object position $(x_j, x_q)$, then add triple $(s, r, o)$ to $T$.
        \ENDIF
        \ENDFOR
			
		\FOR[UL $\leftarrow$ BR]{token pair $(x_i, x_j) \in \mathbf{Y}_r^{\textit{BR}}$}
        \STATE
            Following the same manner as above for ``UL'' finding ``BR'', finally add decoded triple $(s, r, o)$ to $T$.
		\ENDFOR
  
		\ENDFOR
		\RETURN $T$
	\end{algorithmic}
\end{algorithm}

Taking Figure~\ref{fig:TaggingStrategy} as an example, suppose token pairs $(\textit{USA}, \textit{Seattle})$, $(\textit{USA}, \textit{New})$, $(\textit{USA}, \textit{York})$, $(\textit{USA}, \textit{City})$ are filled with ``SP'', ``UL'', ``BR'', ``BR'' respectively. First, we need to find all the single points on the table, then we obtain the entity pair $(\textit{USA}, \textit{Seattle})$. Then, taking the ``UL'' $(\textit{USA}, \textit{New})$ as the anchor point, finding the nearest ``BR'' $(\textit{USA}, \textit{York})$ on the bottom-right, and obtaining the entity pair $(\textit{USA}, \textit{New York})$. Finally, using ``BR'' $(\textit{USA}, \textit{City})$ as the anchor point, find the nearest ``UL'' $(\textit{USA}, \textit{New})$ on the upper-left, and obtain the entity pair $(\textit{USA}, \textit{New York City})$. Thus we decode the entity pairs $(\textit{USA}, \textit{Seattle})$, $(\textit{USA}, \textit{New York})$, $(\textit{USA}, \textit{New York City})$ for the relation ``$\textit{Contains}$''.

\subsection{Loss Function} 

Our training object is to minimize cross entropy loss function:

\begin{equation}
\mathcal{L} = - \frac{1}{N^2 \times |\mathcal{R}|}\sum_{i=1}^{N} \sum_{j=1}^{N} \sum_{r=1}^{|\mathcal{R}|} \log{P_r(\mathbf{y}_{ij}=\mathbf{y}^\prime_{ij}| S)}
\end{equation}

where $\mathbf{y}^\prime_{ij}$ is golden label of relation-specific table under our tagging scheme, $r$ is relation in predefined set of relations $\mathcal{R}$.

\section{Experiments}

\subsection{Experiments Settings}
\noindent \textbf{Datasets}
According to previous work, we evaluate our proposed model by choosing two widely used datasets for relational triple extraction: NYT~\cite{DBLP:conf/pkdd/RiedelYM10} and WebNLG~\cite{gardent2017creating}. Typically there are two different variants, partial matching (denote with $^\star$) and exact matching. Partial matching means that only the last word of each subject and object is annotated, while exact matching strictly annotates the first and last word of each subject and object. However,~\citet{DBLP:conf/naacl/LeeLYY22} points out that evaluating generalization ability of RTE methods with existing datasets is unrealistic. So we still compare our model with other state-of-the-art models in the rearranged dataset NYT-R and WebNLG-R~\cite{DBLP:conf/naacl/LeeLYY22}. We will discuss this further in Section~\ref{sec:generalize}. Statistics of all datasets are shown in Table~\ref{tab:dataset_statistics}.

\begin{table}[htbp]
    \centering
    \renewcommand\arraystretch{1}
    \setlength\tabcolsep{.85 mm}
    \scalebox{0.85}{
        \begin{tabular}{lccrcccc}
        \toprule
        \multicolumn{1}{c}{\multirow{2}[4]{*}{Dataset}} & \multirow{2}[4]{*}{Train} & \multirow{2}[4]{*}{Test} &       & \multicolumn{4}{c}{Overlapping Pattern} \\
        \cmidrule{5-8}      &       &       &       & Normal & SEO   & EPO   & HTO \\
        \midrule
        NYT   & \multirow{3}[2]{*}{56196} & \multirow{3}[2]{*}{5000} &       & 3071  & 1273  & 1168  & 117 \\
        NYT$^\star$ &       &       &       & 3266  & 1297  & 978   & 45 \\
        NYT-R &       &       &       & 3965  & 709   & 408   & 110 \\
        \midrule
        WebNLG & \multirow{3}[2]{*}{5019} & \multirow{3}[2]{*}{703} &       & 239   & 448   & 6     & 85 \\
        WebNLG$^\star$ &       &       &       & 246   & 457   & 26    & 83 \\
        WebNLG-R &       &       &       & 389   & 293   & 9     & 84 \\
        \bottomrule
        \end{tabular}%
    }
    \caption{Dataset Statistics. ``-R'' represents dataset with rearranged variants. Overlapping patterns statistics on test set only. HTO is also referred as SOO in some works~\citet{DBLP:conf/acl/ZhengWCYZZZQMZ20}. Detailed statistics on the rest of the dataset can be found in Appendix~\ref{sec:dataset_details}.}
    \label{tab:dataset_statistics}%
\end{table}%

\noindent \textbf{Evaluation Metrics}
We use three standard evaluation metrics to evaluate how consistently the relational triples are predicted: Micro Precision (Prec.), Recall (Rec.), and F1 Score (F1). In partial matching, triple is considered correct if the predicted relation, tail token of the subject, and tail token of the object are all correct. In exact matching, triple is considered correct if the predicted relation and the entire boundaries of both the subject and object are correct.

\noindent \textbf{Implementation Details}
For a fair comparison with the previous work~\citep{DBLP:conf/acl/LiuZZHZ18,DBLP:conf/acl/FuLM19,DBLP:conf/coling/WangYZLZS20, DBLP:conf/acl/ZhengWCYZZZQMZ20, DBLP:conf/aaai/ShangHM22}, we use bert-base-cased\footnote{More details at https://huggingface.co/bert-base-cased} with almost 109M parameters as the default pre-trained language model, with hidden size $d$ of 768. All parameters are fine-tuned on NYT/WebNLG with learning rates of 4e-5/3e-5, and batchsize of 24/6. We use Adam~\cite{DBLP:journals/corr/KingmaB14} with linear decay of learning rate as optimizer. Consistent with previous work, we set the maximum sentence length for training to 100. All our experiments were conducted on one NVIDIA RTX A6000 GPU.

\subsection{Main Results}

We took twelve powerful state-of-the-art RTE methods as baselines to compare with our  model: CopyRE~\cite{DBLP:conf/acl/LiuZZHZ18}, GraphRel~\cite{DBLP:conf/acl/FuLM19}, RSAN~\cite{DBLP:conf/ijcai/YuanZPZSG20}, MHSA~\cite{ijcai2020p524}, CasRel~\cite{DBLP:conf/acl/WeiSWTC20}, TPLinker~\cite{DBLP:conf/coling/WangYZLZS20}, SPN~\cite{DBLP:journals/corr/abs-2011-01675}, PRGC~\cite{DBLP:conf/acl/ZhengWCYZZZQMZ20}, TDEER~\cite{DBLP:conf/emnlp/LiLDYLH21}, EmRel~\cite{DBLP:conf/naacl/XuWLSZGM22}, OneRel~\cite{DBLP:conf/aaai/ShangHM22}, QIDN~\cite{DBLP:conf/emnlp/Tan0HZC0Z22}. For fair comparison, all results are taken from the original paper.

The main experimental results in Table~\ref{tab:main_results} demonstrate the effectiveness of our proposed method. In total, our method outperforms twelve baselines in almost all metrics, and achieves the new state-of-the-art. Specifically, RTF achieves the best performance among all table filling approaches. RTF improved +0.4\%, +0.6\% over OneRel on NYT and WebNLG, and +0.5\%, +0.3\% on NYT$^\star$ and WebNLG$^\star$. These improvements may come from fully leveraging regional information.

RTF improves performance significantly for exact matching. In particular, RTF achieves a +0.6\% improvement on WebNLG with 216 relations. This indicates RTF fully exploits the local dependencies of triples, instead of using a shallow representation of token pair-level. RTF can also gain from determining the entity pair boundaries which are learned from massive relations. The improvement of RTF in partial matching is lower than that in exact matching. Because NYT$^\star$ and WebNLG$^\star$ only need to identify the last token of the subject and object, which leads to a smaller gain of RTF from regional information. Therefore, the full potential of RTF cannot be fully utilized for partial matching. It is worth to note that RTF has a remarkable improvement in recall on all datasets. We will further explore this phenomenon in Section~\ref{sec:local_depen}.

\begin{table*}[htbp]
    \centering
    \renewcommand\arraystretch{1}
    \setlength\tabcolsep{.9 mm}
    \scalebox{0.98}{
    \begin{tabular}{lccccccccccccccc}
    \toprule
    \multicolumn{1}{c}{\multirow{2}[4]{*}{Model}} & \multicolumn{3}{c}{NYT$^\star$} &       & \multicolumn{3}{c}{NYT} &       & \multicolumn{3}{c}{WebNLG$^\star$} &       & \multicolumn{3}{c}{WebNLG} \\
    \cmidrule{2-4}\cmidrule{6-8}\cmidrule{10-12}\cmidrule{14-16}      & Prec. & Rec.  & F1    &       & Prec. & Rec.  & F1    &       & Prec. & Rec.  & F1    &       & Prec. & Rec.  & F1 \\
    \midrule
    CopyRE~\cite{DBLP:conf/acl/LiuZZHZ18} & 61.0  & 56.6  & 58.7  &       & -     & -     & -     &       & 37.7  & 36.4  & 37.1  &       & -     & -     & - \\
    GraphRel~\cite{DBLP:conf/acl/FuLM19} & 63.9  & 60.0  & 61.9  &       & -     & -     & -     &       & 44.7  & 41.1  & 42.9  &       & -     & -     & - \\
    RSAN~\cite{DBLP:conf/ijcai/YuanZPZSG20} & -     & -     & -     &       & 85.7  & 83.6  & 84.6  &       & -     & -     & -     &       & 80.5  & 83.8  & 82.1  \\
    MHSA~\cite{ijcai2020p524} & 88.1  & 78.5  & 83.0  &       & -     & -     & -     &       & 89.5  & 86.0  & 87.7  &       & -     & -     & - \\
    \midrule
    CasRel~\cite{DBLP:conf/acl/WeiSWTC20} & 89.7  & 89.5  & 89.6  &       & -     & -     & -     &       & 93.4  & 90.1  & 91.8  &       & -     & -     & - \\
    TPLinker~\cite{DBLP:conf/coling/WangYZLZS20} & 91.3  & 92.5  & 91.9  &       & 91.4  & 92.6  & 92.0  &       & 91.8  & 92.0  & 91.9  &       & 88.9  & 84.5  & 86.7  \\
    SPN~\cite{DBLP:journals/corr/abs-2011-01675} & \underline{93.3}  & 91.7  & 92.5  &       & 92.5  & 92.2  & 92.3  &       & 93.1  & 93.6  & 93.4  &       & -     & -     & - \\
    PRGC~\cite{DBLP:conf/acl/ZhengWCYZZZQMZ20} & \underline{93.3}  & 91.9  & 92.6  &       & \underline{93.5}  & 91.9  & 92.7  &       & 94.0  & 92.1  & 93.0  &       & 89.9  & 87.2  & 88.5  \\
    TDEER~\cite{DBLP:conf/emnlp/LiLDYLH21} & 93.0  & 92.1  & 92.5  &       & -     & -     & -     &       & 93.8  & 92.4  & 93.1  &       & -     & -     & - \\
    EmRel~\cite{DBLP:conf/naacl/XuWLSZGM22} & 91.7  & 92.5  & 92.1  &       & 92.6  & 92.7  & 92.6  &       & 92.7  & 93.0  & 92.9  &       & 90.2  & 87.4  & 88.7  \\
    OneRel~\cite{DBLP:conf/aaai/ShangHM22} & 92.8  & 92.9  & 92.8  &       & 93.2  & 92.6  & 92.9  &       & \underline{94.1}  & 94.4  & 94.3  &       & 91.8  & 90.3  & 91.0  \\
    QIDN~\cite{DBLP:conf/emnlp/Tan0HZC0Z22} & \underline{93.3}  & 92.5  & 92.9  &       & 93.4  & 92.6  & 93.0  &       & \underline{94.1}  & 93.7  & 93.9  &       & -     & -     & - \\
    \midrule
    RTF   & 93.2  & \textbf{93.4} & \textbf{93.3} &       & \underline{93.5}  & \textbf{93.2} & \textbf{93.3} &       & 94.0  & \textbf{95.3} & \textbf{94.6} &       & \textbf{92.0} & \textbf{91.1} & \textbf{91.6} \\
    \bottomrule
    \end{tabular}%
    }
  \caption{Main results. The \underline{underline} means they are both the maximum.}
  \label{tab:main_results}%
\end{table*}%

\subsection{Analysis on Complex Scenarios}

According to previous works~\citep{DBLP:conf/acl/WeiSWTC20, DBLP:conf/coling/WangYZLZS20, DBLP:journals/corr/abs-2011-01675, DBLP:conf/acl/ZhengWCYZZZQMZ20, DBLP:conf/emnlp/LiLDYLH21, DBLP:conf/aaai/ShangHM22}, we also evaluated our model under different overlapping triples and different number of triples. 

The results in Table~\ref{tab:complex_scenarios} show that our model works effectively in complex scenarios. RTF is comparable to OneRel in handling different overlapping patterns in NYT$^\star$, and outperforms other methods in most scenarios with different numbers of triples. RTF outperforms OneRel by +1.0\% and +1.1\% for T=1 and T=3, respectively. OneRel better than RTF for T$\geq$5. We attribute to OneRel increasing the number of valid tags by doubling the sentence length. We disregard this because it would increase memory consumption significantly. In Normal subset, RTF achieves +0.8\% and +0.9\% improvement over OneRel for NYT$^\star$ and WebNLG$^\star$, respectively, significantly outperforming other methods. 


In addition, we observe that RTF performs better on WebNLG$^\star$ than on NYT$^\star$, especially in EPO where it achieves +0.7\% improvement over OneRel. We argue that WebNLG$^\star$ has more types of relations, thus there is more entity pair boundary information to share, and more regional dependencies to exploit. 

\subsection{Analysis on Generalization Capability}\label{sec:generalize}

Although the current RTE approach has been successful on  widely used datasets, however, this performance may be unrealistic. Therefore, we evaluate generalization capability of our method on NYT-R and WebNLG-R~\cite{DBLP:conf/naacl/LeeLYY22}. Same as previous settings~\cite{DBLP:conf/naacl/LeeLYY22}, we trained RTF and OneRel with 300 and 500 epochs on NYT-R and WebNLG-R respectively. 

\begin{table}[t]
    \centering
    \setlength\tabcolsep{1. mm}
    \renewcommand\arraystretch{1}
    \begin{tabular}{lcccrccc}
    \toprule
    \multicolumn{1}{c}{\multirow{2}[4]{*}{Model}} & \multicolumn{3}{c}{NYT-R} &       & \multicolumn{3}{c}{WebNLG-R} \\
\cmidrule{2-4}\cmidrule{6-8}          & Prec. & Rec.  & F1    &       & Prec. & Rec.  & F1 \\
    \midrule
    CasRel & 65.9  & 60.1  & 62.9  &       & 73.6  & 64.2  & 68.6  \\
    TPLinker & \textbf{69.0}  & 60.8  & 64.7  &       & 75.1  & 63.9  & 69.1  \\
    PRGC  & 63.5  & 61.6  & 62.6  &       & 61.6  & 62.0  & 61.8  \\
    OneRel$^\dagger$ & 67.3  & 60.8  & 63.9  &       & 75.2  & 65.7  & 70.1  \\
    \midrule
    RTF    & 68.8 & \textbf{62.3} & \textbf{65.4} &       & \textbf{76.5} & \textbf{66.5} & \textbf{71.2} \\
    \bottomrule
    \end{tabular}%
    \caption{Generalization capability evaluation. $\dagger$ means not reported in the original paper, results are obtained by running the provided source code.}
  \label{tab:generalization_study}%
\end{table}%

Experimental results in Table~\ref{tab:generalization_study} show that although OneRel achieves state-of-the-art on widely used datasets, it still cannot outperform TPLinker on NYT-R. This means OneRel has a strong overfitting ability on widely used datasets and prefers memorization to generalization. In contrast, RTF shows consistent improvement on NYT-R and WebNLG-R and displays stronger generalization ability than TPLinker and OneRel. This indicates that our method can effectively generalize to unseen triples. Such capability may be derived from relational residual learning and a more robust decoding strategy. In addition, we can observe from the results of Table~\ref{tab:generalization_study} that there is a significant difference in the performance of the models between the widely used datasets and the rearranged datasets. Therefore, we also advocate evaluating the generalization capability of RTE methods.

\begin{table*}[htbp]
    \centering
    \setlength\tabcolsep{.75 mm}
    \renewcommand\arraystretch{1}
    \scalebox{0.85}{
    \begin{tabular}{lccccccccccccccccccc}
    \toprule
    \multicolumn{1}{c}{\multirow{2}[4]{*}{Model}} & \multicolumn{9}{c}{NYT$^\star$}                                       &       & \multicolumn{9}{c}{WebNLG$^\star$} \\
    \cmidrule{2-10}\cmidrule{12-20}      & Normal & EPO   & SEO   & HTO   & T=1   & T=2   & T=3   & T=4   & T$\geq$5 &       & Normal & EPO   & SEO   & HTO   & T=1   & T=2   & T=3   & T=4   & T$\geq$5 \\
    \midrule
    CasRel & 87.3  & 92.0  & 91.4  & 77.0$^\dagger$ & 88.2  & 90.3  & 91.9  & 94.2  & 83.7  &       & 89.4  & 94.7  & 92.2  & 90.4  & 89.3  & 90.8  & 94.2  & 92.4  & 90.9  \\
    TPLinker & 90.1  & 94.0  & 93.4  & 90.1$^\dagger$ & 90.0  & 92.8  & 93.1  & 96.1  & 90.0  &       & 87.9  & 95.3  & 92.5  & 86.0  & 88.0  & 90.1  & 94.6  & 93.3  & 91.6  \\
    SPN   & 90.8  & 94.1  & 94.0  & -     & 90.9  & 93.4  & 94.2  & 95.5  & 90.6  &       & -     & -     & -     & -     & 89.5  & 91.3  & 96.4  & 94.7  & 93.8  \\
    PRGC  & 91.0  & 94.5  & 94.0  & 81.8  & 91.1  & 93.0  & 93.5  & 95.5  & 93.0  &       & 90.4  & 95.9  & 93.6  & 94.6  & 89.9  & 91.6  & 95.0  & 94.8  & 92.8  \\
    TDEER & 90.8  & 94.5  & 94.1  & -     & 90.8  & 92.8  & 94.1  & 95.9  & 92.8  &       & 90.7  & 95.4  & 93.5  & -     & 90.5  & 93.2  & 94.6  & 93.8  & 92.3  \\
    OneRel & 90.6  & \textbf{95.1} & 94.8  & \textbf{90.8} & 90.5  & 93.4  & 93.9  & \textbf{96.5} & \textbf{94.2} &       & 91.9  & 95.4  & 94.7  & \textbf{94.9} & 91.4  & \textbf{93.0} & 95.9  & 95.7  & 94.5  \\
    \midrule
    RTF   & \textbf{91.4} & 94.9  & \textbf{94.9} & 88.3  & \textbf{91.5} & \textbf{93.7} & \textbf{95.0} & 96.3  & 93.1  &       & \textbf{92.8} & \textbf{96.1} & \textbf{95.0} & 94.8  & \textbf{92.4} & 92.9  & \textbf{96.4} & \textbf{95.8} & \textbf{95.0} \\
    \bottomrule
    \end{tabular}%
    }
    \caption{F1-score on sentences with different overlapping patterns and different triple numbers. T is the number of triples contained in the sentence. $\dagger$ means the results reported by \citet{DBLP:conf/acl/ZhengWCYZZZQMZ20}.}
    
  \label{tab:complex_scenarios}%
\end{table*}%

\subsection{Analysis on Regional Correlation}\label{sec:local_depen}

In Table~\ref{tab:main_results}, we observe that almost all improvements in F1 of RTF derive from recall. RTF improves the recall on NYT$^\star$, NYT, WebNLG$^\star$, WebNLG by +0.5\%, +0.6\%, +0.9\%, +0.8\% over OneRel, respectively. 

\begin{figure}[ht]
    \centering
    \includegraphics[width=0.48\textwidth]{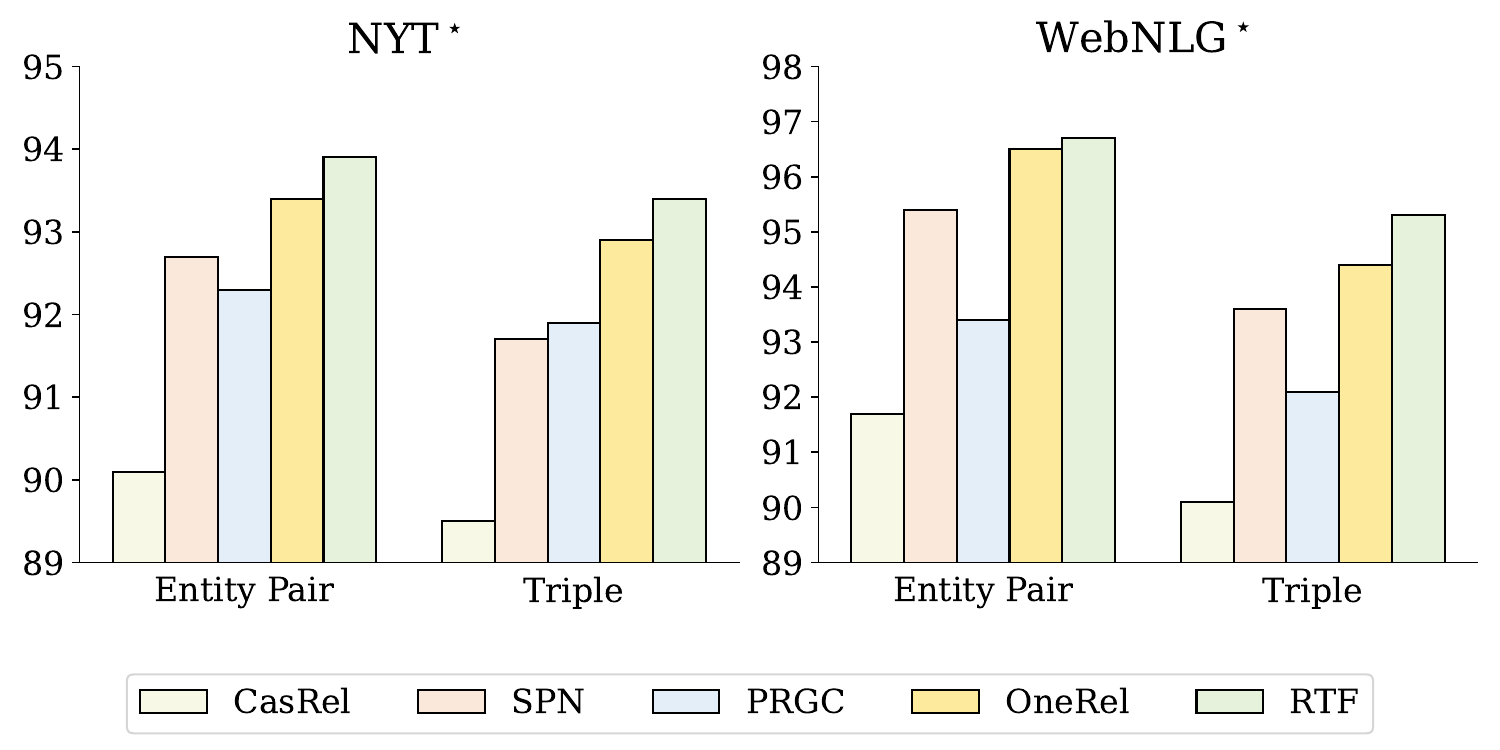}
    \caption{Recall for two different tasks, entity pair recognition and triple extraction.}
    \label{fig:regional_corr}
\end{figure}

We assume RTF improves triple extraction by detecting more entity pairs with regional correlation. 
Figure~\ref{fig:regional_corr} results support this hypothesis. RTF recalls more entity pairs $(h, t)$ than OneRel on NYT$^\star$, and thus achieves better recall of the triples $(h, r, t)$. This implies that introducing regional correlation of triples can detect more entity pairs, which is highly beneficial for RTE. It appears that the improvement in recall of RTF on WebNLG$^\star$ is not substantial. We attribute this to the presence of a large number of overlapping triple patterns in RTE, especially multi-relation dataset. Overlapping leads to the same entity or entity pair existing in multiple triples. Although RTF only detects slightly more entity pairs, it significantly improves the recall of triples on WebNLG$^\star$.

\begin{table}[ht]
    \centering
    \setlength\tabcolsep{.75 mm}
    \renewcommand\arraystretch{1}
    \begin{tabular}{lccccccc}
    \toprule
    \multicolumn{1}{c}{\multirow{2}[4]{*}{Model}} & \multicolumn{3}{c}{NYT} &       & \multicolumn{3}{c}{WebNLG} \\
    \cmidrule{2-4}\cmidrule{6-8}      & Prec. & Rec.  & F1    &       & Prec. & Rec.  & F1 \\
    \midrule
    RTF   & \textbf{93.5} & \textbf{93.2} & \textbf{93.3} &       & \textbf{92.0} & \textbf{91.1} & \textbf{91.6} \\
    \midrule
    w/o CB & 93.0  & 92.7  & 92.9  &       & 91.6  & 89.9  & 90.7  \\
    w/o RR & 93.1  & 92.8  & 93.0  &       & 90.6  & 90.2  & 90.4  \\
    w/o CB, RR & 92.9  & 92.5  & 92.7  &       & 90.9  & 89.5  & 90.2  \\
    \bottomrule
    \end{tabular}%
    
    \caption{Ablation Study. ``CB'' means Convolution Block, ``RR'' means Relational Residual Learning. }
  \label{tab:ablation}%
\end{table}%

To further investigate the effectiveness of the two modules of our method that utilize region correlations, we conduct ablation experiments next. From Table~\ref{tab:ablation}, we can observe that when either Convolution Block (CB) or Relational Reisdual (RR) is removed, there is a different degree of performance degradation in our models on both NYT and WebNLG. Removing both simultaneously will result in more noticeable decreases. This illustrates the effectiveness of both. In particular, removing convolution blocks or relation residuals in WebNLG degrades performance dramatically. Since there are numerous relations in WebNLG, this hurts shared entity pair boundary information across relations. In addition, we observe a significant performance degradation when removing convolution block and the relational residual concurrently (-0.6\% on NYT and -1.4\% on WebNLG), which suggests that they are complementary to each other and utilize local dependencies more adequately when combined.

\section{Conclusion}

In this study, we propose a novel region-based table filling method to fully exploit the local spatial correlations of relational triples. We devise a novel region-based tagging scheme and decoding algorithm to identify triples by determining two endpoints of each region. We use convolution to capture regional correlations of triples on a table, and reduce learning stress for relation classifiers by sharing the tagging score between different relations. Experimental results demonstrate that RTF achieves state-of-the-art performance on two benchmark datasets with better generalization ability.

\clearpage
\newpage

\section*{Limitations}
Despite the superior performance of table filling-based methods, they have some disadvantages from the perspective of resource usage, such as larger GPU memory consumption and requiring longer training time. 

For example, due to the limitation of two-dimensional table structure, when the sentence length increases, the memory consumption increases exponentially and the speed becomes slower due to the expansion of the table area.  Alternatively, when the number of relations increases, more cells need to be filled as the number of relations increases. The mentioned problem is not serious in one-dimensional sequence-based labeling approaches.

Therefore, the problem becomes difficult when extending from sentence-level relational triple extraction to document-level relational triple extraction. 


\bibliography{anthology,custom}
\bibliographystyle{acl_natbib}

\clearpage
\newpage
\appendix



\section{Detailed dataset statistics}
\label{sec:dataset_details}

The remaining detailed statistics for all variants of NYT and WebNLG are shown in Table~\ref{tab:detailed_basic} and Table~\ref{tab:detailed_num_triples}. NYT has 24 relations, while WebNLG has far more relations than NYT. Avg.Len represents the average length of entity tokens, and Avg.Area represents the average area of the triple on the table. Note that the area of all triples are almost always around 9, just covered by one $\text{Conv}3\times3$.

\begin{table}[ht]
    \centering
	\small
	\setlength{\tabcolsep}{1.mm}
    \renewcommand\arraystretch{1}
    \begin{tabular}{lccc}
    \toprule
    \multicolumn{1}{c}{Dataset} & Relations & Avg. Len & Avg. Area \\
    \midrule
    NYT   & 24    & 1.83  & 3.25  \\
    NYT$^\star$ & 24    & 1.33  & 1.71  \\
    NYT-R & 24    & 2.40  & 5.11  \\
    \midrule
    WebNLG & 216   & 3.30  & 10.84  \\
    WebNLG$^\star$ & 171   & 1.56  & 2.43  \\
    WebNLG-R & 216   & 3.46  & 12.66  \\
    \bottomrule
    \end{tabular}%
    \caption{Statistics on the number of relations, the average length of entities, and the average area of triads. Avg. is an abbreviation for Average. }
    \label{tab:detailed_basic}
\end{table}

\begin{table}[ht]
    \centering
	\small
	\setlength{\tabcolsep}{1.mm}
    \renewcommand\arraystretch{1}
    \begin{tabular}{lccccc}
    \toprule
    \multicolumn{1}{c}{Dataset} & T=1   & T=2   & T=3   & T=4   & T$\geq$5 \\
    \midrule
    NYT   & 3089  & 1137  & 300   & 317   & 157 \\
    NYT$^\star$ & 3244  & 1045  & 312   & 291   & 108 \\
    NYT-R & 3958  & 733   & 192   & 50    & 67 \\
    \midrule
    WebNLG & 256   & 175   & 138   & 93    & 41 \\
    WebNLG$^\star$ & 266   & 171   & 131   & 90    & 45 \\
    WebNLG-R & 411   & 143   & 91    & 59    & 10 \\
    \bottomrule
    \end{tabular}%
    \caption{Statistics for the number of different triples in each sentence in the test set. }
    \label{tab:detailed_num_triples}
\end{table}

\section{Interpretation of Convolutional Interaction}\label{sec:conv_interact}

\begin{figure*}[t!]
    \centering
    \scalebox{0.8}{
    \includegraphics[width=.9\textwidth]{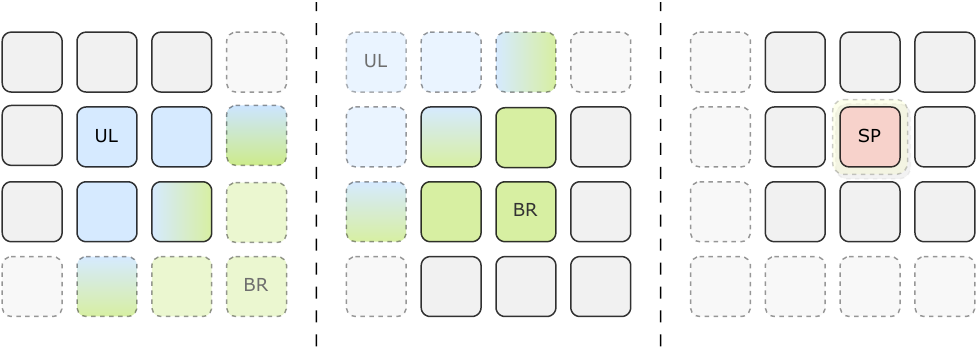}
    }
    \caption{Interaction when combining convolution with EPR tagging scheme. From left to right is the scene when the convolution interacts with ``UL'', ``BR'' and ``SP'' respectively.}
    \label{fig:conv_interact}
\end{figure*}

Figure~\ref{fig:conv_interact} explains the interaction process when convolution is combined with EPR tagging scheme. 

As the convolution kernel sweeps to the upper-left of the ``UL'', the gray cells are located outside the entity pair boundary and the blue area is located inside. It is easier to detect the location of ``UL'' if the semantics inside the boundary and outside the boundary are different. When the entity pair collapses to a single point, the region information is no longer explicitly in effect, but degenerates to a token level or token pair-level representation.

\section{Visualization}

\begin{figure*}[t!]
    \centering
    \scalebox{0.88}{
    \includegraphics[width=.95\textwidth]{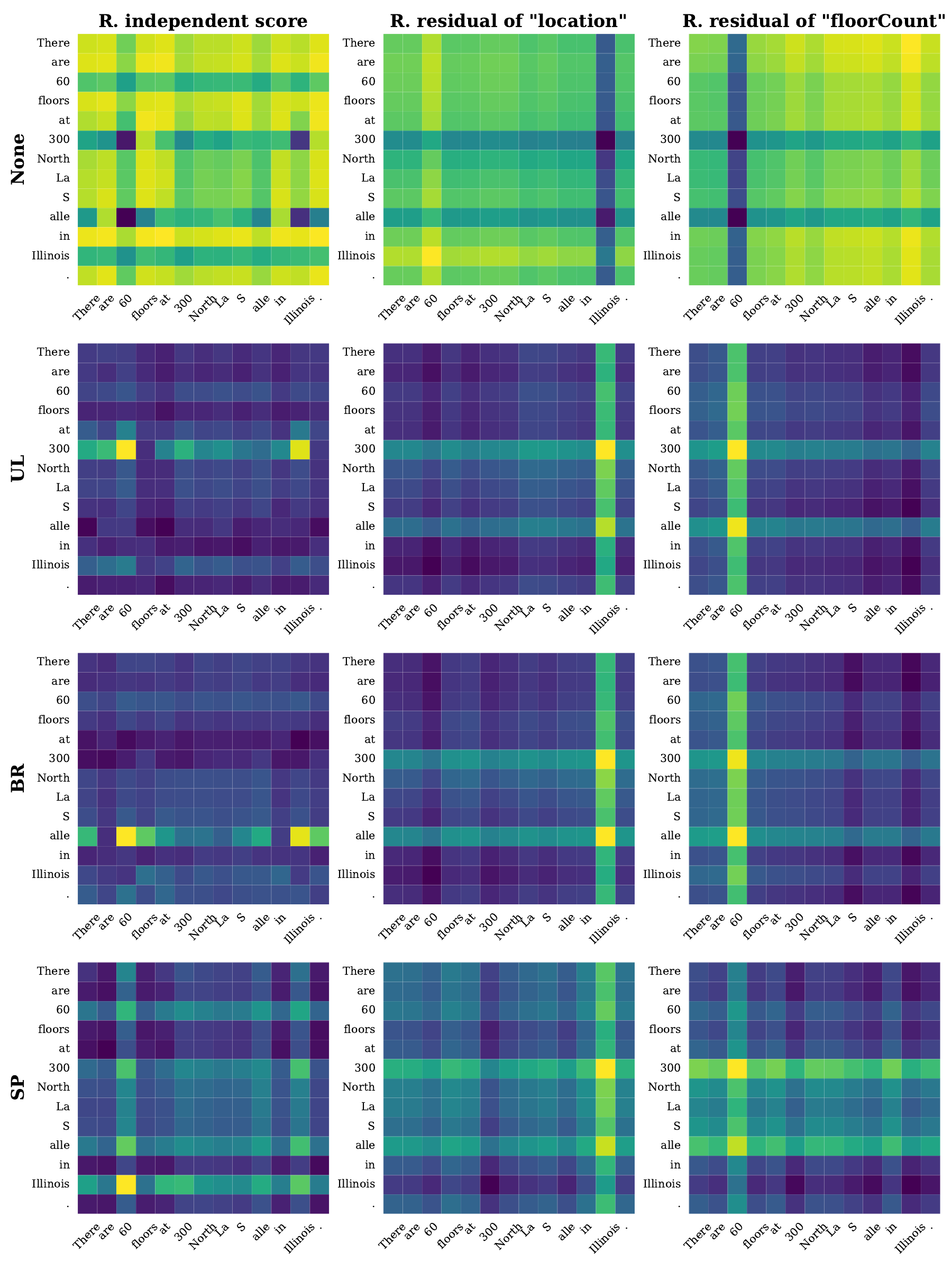}
    }
    \caption{Visualization of Tagging Score. Brighter colors mean larger scores. Each column from left to right is the relation independent score $\boldsymbol{z}_{ij}^{e}$, the relation residual $\boldsymbol{z}_{ij}^{r}$ for ``\textit{location}'' and ``\textit{floorCount}'', respectively.}
    \label{fig:visualize}
\end{figure*}

Figure~\ref{fig:visualize} shows how RTF obtains each tagging score. For the input sentence ``\textit{There are 60 floors at 300 North LaSalle in Illinois.}'', there are two SEO relational triples (\textit{300 North LaSalle}, \textrm{location}, \textit{Illinois}), (\textit{300 North LaSalle}, \textrm{floorCount}, \textit{60}). 

First, we can observe the significant difference in the working patterns of $\boldsymbol{z}_{ij}^{e}$ and $\boldsymbol{z}_{ij}^{r}$. No matter for which valid tag (UL, BR, SP), the highlighted part of $\boldsymbol{z}_{ij}^{e}$ is always concentrated in a two-dimensional region. In contrast, the highlighted areas of $\boldsymbol{z}_{ij}^{r}$ are always concentrated on rows and columns. This indicates that the Convolution Block utilizes region-level features. 

Moreover, observe that the relation-independent score $\boldsymbol{z}_{ij}^{e}$ always accurately predicts the entity pair boundaries of all triples. For example, in tag ``UL'', $\boldsymbol{z}_{ij}^{e}$ accurately detects that the token pair (\textit{300}, \textit{60}), (\textit{300}, \textit{Illinois}) must be the upper-left corner of the entity pair, regardless of whichever relationship they belong to. This means that $\boldsymbol{z}_{ij}^{e}$ is a more general score shared by all kinds of relations, thus reducing the learning burden of the relation classifier.

Finally, different relational classifiers learn different scores $\boldsymbol{z}_{ij}^{r}$ for the same tag, just complementary to $\boldsymbol{z}_{ij}^{e}$. For example, in tag ``BR'', the classifier with the relation ``$\textit{location}$'' gives high scores for both (\textit{300}, \textit{Illinois}) and (\textit{alle}, \textit{Illinois}), but since (\textit{alle}, \textit{Illinois}) has a higher score in $\boldsymbol{z}_{ij}^{e}$, so (\textit{300}, \textit{Illinois}) is not really the bottom-right corner of the entity pair. This means that $\boldsymbol{z}_{ij}^{e}, \boldsymbol{z}_{ij}^{r}$ are working by combination.

\end{document}